\documentclass[11pt,a4paper]{article}
\usepackage[hyperref]{acl2020}
\usepackage{times}
\usepackage{latexsym}
\usepackage{bm}
\usepackage{array}
\usepackage{siunitx}

\usepackage{amssymb}
\usepackage{amsmath}

\usepackage[pdftex]{graphicx}

\usepackage{microtype}

 \aclfinalcopy 
\def\aclpaperid{1329} 

\newcommand{\sidx}{^{(s)}}
\newcommand{\gold}{^{(s)\ast}}

\title{Using Context in Neural Machine Translation Training Objectives}
  
\author{Danielle Saunders \and Felix Stahlberg\thanks{ \hspace*{0.5em}Now at Google} \and Bill Byrne \\
    Department of Engineering, University of Cambridge, UK  \\
      {\tt ds636@cam.ac.uk\hspace*{0.5em} fstahlberg@google.com
     \hspace*{0.5em} wjb31@cam.ac.uk}}

\begin{document}
\maketitle
\begin{abstract}
We present Neural Machine Translation (NMT) training using document-level metrics with batch-level documents. Previous sequence-objective approaches to NMT training focus exclusively on sentence-level metrics like sentence BLEU which do not correspond to the desired evaluation metric, typically document BLEU. Meanwhile research into document-level NMT training focuses on data or model architecture rather than training procedure. We find that each of these lines of research has a clear space in it for the other, and propose merging them with a scheme that allows a document-level evaluation metric to be used in the NMT training objective.

We first sample pseudo-documents from sentence samples. We then approximate the expected document BLEU gradient with Monte Carlo sampling for use as a cost function in Minimum Risk Training (MRT). This two-level sampling procedure gives NMT performance gains over sequence MRT and maximum-likelihood training. We demonstrate that training is more robust for document-level metrics than with sequence metrics. We further demonstrate improvements on NMT with TER and Grammatical Error Correction (GEC) using GLEU, both metrics used at the document level for evaluations.

\end{abstract}

\section{Introduction}
Neural Machine Translation (NMT) research has explored token-level likelihood functions \cite{sutskever2014sequence, bahdanau2015neural} and sequence-level objectives inspired by reinforcement learning \cite{ranzanto16sequencelevel, bahdanau2016actor} or expected Minimum Risk Training (MRT) \cite{shen2016minimum}. A typical sequence objective in these cases is based on sentence-level BLEU (sBLEU) \cite{edunov2018classical}. However sBLEU, even if aggregated over sentences, is only an approximation of the desired metric, document-level BLEU. Beyond translation, many metrics for natural language tasks do not have robust sentence-level approximations. A logical progression is the extension of sequence-level NMT training objectives to include context from outside the sentence.

Document-based NMT, by contrast, aims to use out-of-sentence context to improve translation. Recent research explores lexical consistency by providing additional sentences during training \cite{maruf-etal-2019-selective, voita-etal-2018-context, voita-etal-2019-context} or inference \cite{voita-etal-2019-context, stahlberg2019neural}, potentially with adjustments to model architecture. However, to the best of our knowledge, no attempt has been made to extend sequence-level neural training objectives to include document-level reward functions. This is despite document-level BLEU being arguably the most common NMT metric, and being the function originally optimised by  Minimum Error Rate Training (MERT) for Statistical Machine Translation (SMT) \cite{och2003minimum}.

We propose merging lines of research on training objectives and document-level translation. We achieve this by presenting a document-level approach to sequence-level objectives which brings the training objective closer to the actual evaluation metric, using MRT as a representative example. We demonstrate MRT under document-level BLEU as well as Translation Edit Rate (TER) \cite{snover2006study}, which while decomposable to sentence level is less noisy when used over documents. We consider both pseudo-documents where sentences are assigned randomly to a mini-batch, and true document context where all sentences in the batch are from the same document. 

We finally apply our scheme to supervised Grammatical Error Correction, for which using neural models is becoming increasingly popular \citep{neural-first,neural-rl,stahlberg2019neural}. 
We show gains in GEC metrics GLEU \cite{gleu} and M2 \cite{m2}.

\subsection{Related Work}
Minimum Error Rate Training was introduced for phrase-based SMT with document-level BLEU \cite{och2003minimum}. \citet{shen2016minimum} extend these ideas to NMT, using expected minimum risk at the sequence level with an sBLEU cost for end-to-end NMT training.  \citet{edunov2018classical} explore random and beam sampling for NMT sequence-MRT, as well as other sequence-level training losses. 

Related developments in NMT include combined reinforcement-learning/cross-entropy approaches such as MIXER \cite{ranzanto16sequencelevel}, which itself has origins in the REINFORCE algorithm described by \citet{williams1992simple}. We do not explore such approaches, although our document-sampling and document-metric schemes could in principle be extended to them.

Sequence-level MRT has seen success outside NMT. \citet{ayana2016neural} use sequence MRT for summarization, while \citet{shannon2017optimizing} uses a related approach for speech recognition.
MRT can be seen as a special case of neural reinforcement learning, which \citet{neural-rl} apply to GEC with sequence-level costs. Closest to our approach is the work of \citet{jean2019context} on NMT with a minibatch-context-sensitive training procedure. However, they do not optimize on document metrics over those contexts. They also sample contexts randomly, while we find diverse context sampling is important for the success of document-MRT.

\section{Background}

\subsection{Sequence-level MRT}

Sentence-level MRT for NMT aims to minimize the expected loss on training data with a loss function between sampled target sentences $\bm{y}$ and gold reference sentences $\bm{y}^\ast$. For NMT a common sentence-level cost function $\Delta(\bm{y},\bm{y}^\ast)$ is 1 - sBLEU, where sBLEU is smoothed by setting initial n-gram counts to 1 \cite{edunov2018classical}. 

We take $N$ samples for each of the $S$ sentences in a mini-batch. We write the cost function between the $s^{th}$ reference in a mini-batch, $\bm{y}\gold$, and its $n^{th}$ sample,   $\bm{y}_n\sidx$, as $\Delta_n\sidx =  \Delta(\bm{y}_n\sidx, \bm{y}\gold)$. 
The risk gradient for end-to-end NMT with MRT as in \citet{shen2016minimum}, with sample-count scaling, is then:
\begin{equation}
\nabla_\theta R(\theta) = \frac{1}{N}\sum_{s=1}^S \sum_{n=1}^N\Delta_n\sidx  \frac{\partial}{\partial \theta} \log P(\bm{y}_n\sidx|\bm{x}\sidx;\theta)\label{eq:shen}
\end{equation}

\subsection{Document-level MRT}
By analogy with sequence-level MRT, we consider MRT over batches of $S$ sentence pairs, which we treat as a pseudo-document. In practice we experiment both with sentences chosen randomly from all training data, and with true context where all sentences per batch are from a single document.

Let $X = [\bm{x}^{(1)}, \ldots, \bm{x}^{(S)}]$ be the source document,  $Y = [\bm{y}^{(1)}, \ldots, \bm{y}^{(S)}]$ be a document of candidate translations, and $Y^\ast = [\bm{y}^{(1)\ast}, \ldots, \bm{y}^{(S)\ast}]$ be the reference translations. Document-level metric $D(Y, Y^*)$, which may be non-differentiable, replaces the sequence-level metric $\Delta(\bm{y}, \bm{y}\gold)$. We define the document-level risk:
$$    R(\theta) = \sum_Y D(Y, Y^*) P(Y|X;\theta)$$

Using $p_\theta\nabla_\theta \log p_\theta = \nabla p_\theta$:
\begin{align}
\nabla_\theta R(\theta) &= \sum_Y D(Y, Y^*) P(Y|X;\theta) \nabla_\theta \log P(Y|X;\theta) \notag\\
 &= \mathbb{E}\big[ D(Y, Y^*) \nabla_\theta \log P(Y|X;\theta) |X;\theta\big]
\end{align}

Using simple Monte-Carlo, after \citet{shannon2017optimizing}, we replace the expectation by an average taken over $N$ sampled translation documents $Y_n \sim P(Y|X;\theta)$
$$
\nabla_\theta R(\theta) \approx \frac{1}{N}\sum_{n=1}^N D(Y_n, Y^*) \nabla_\theta \log P(Y_n|X;\theta)
$$
The $n^{th}$ sample for the $s^{th}$ sentence in the batch-level document, $\bm{y}_n\sidx$, contributes the following to the overall gradient:
$$
\nabla_\theta R(\theta)  \approx
\frac{1}{N}\hspace*{-2ex} \sum_{Y: \bm{y}^{(s)} = \bm{y}^{(s)}_n} D(Y, Y^*)   \nabla_\theta  \log P(\bm{y}^{(s)}_n|\bm{x}^{(s)};\theta)  
$$
In other words the gradient of each sample is weighted by the aggregated document-level scores for documents in which the sample appears. 

\subsection{Mini-batch level document sampling}
\begin{figure}[ht!]
\centering
\small
\includegraphics[width=1.02\linewidth]{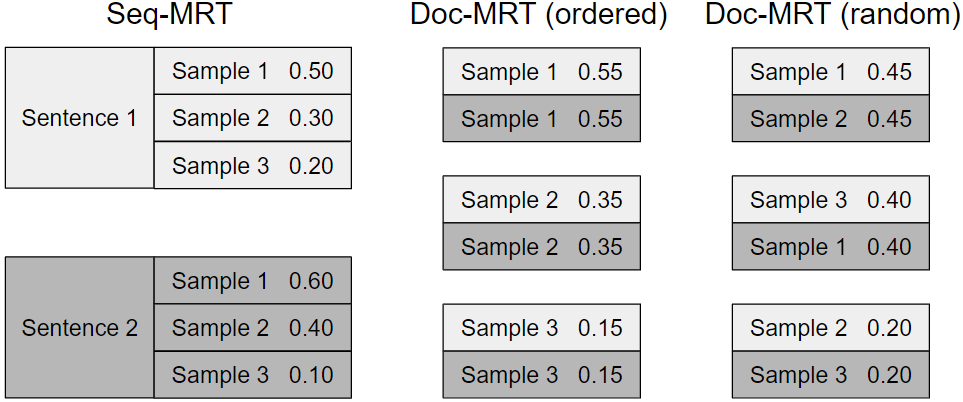}
\caption{Sample-ordering schemes for MRT with $S=2$ sentences / batch and $N=3$ samples / sentence, showing sample costs. In sequence-MRT each sample has its own cost (e.g. sBLEU). For doc-MRT (ordered), samples are ordered and sorted into N-wise `documents', each with a combined cost (e.g. document BLEU). The ordered assignment enforces an extreme range of combined costs. In doc-MRT (random), samples are randomly assigned, making documents on average less diverse with less distinct scores, with a low likelihood of extreme distributions.}
\label{fig:ordering-schemes}

\end{figure}

\label{sec:doc-sample}
To generate sample documents we first sample sentences. Sentence sampling for NMT generates new tokens in a left-to-right manner  \cite{shen2016minimum}. In left-to-right generation each token is sampled from a distribution conditioned on previously sampled tokens, minimizing exposure bias to gold references which the model is unlikely to see at inference time \cite{ranzanto16sequencelevel}. Sampling can be via beam search, or random sampling from the model distribution given previously sampled tokens. Beam search produces more likely samples which may be less diverse compared to random sampling \cite{edunov2018classical}. 

Here we only consider sampling during training. While samples can be more easily generated offline with respect to fixed model parameters, such samples are not representative of the current model.

With $N$ sample translations for each of the $S$ sentence pairs per batch we can construct $N^S$ possible sample documents as sequences of $S$ sentences. Considering all possible documents is intractable unless $N$ and $S$ are small. It also carries the risk that a single sentence will appear in multiple sampled documents, giving it undue weight.

Instead we propose creating $N$ documents by first ordering samples for each sentence (e.g. by sBLEU), then creating the $n^{th}$ sample document $Y_n$ by concatenating the $n^{th}$ sample from each sentence.
This gives a set of $N$ diverse documents sampled from $N^S$ possibilities. We expect the sampled documents to be diverse in contents, since a given sentence will only ever occur in a single document context, and diverse in score. We refer to this scheme as ordered document sampling.

Figure \ref{fig:ordering-schemes} illustrates ordered document sampling by comparison to a scheme which randomly samples sentences to form documents.

\section{Experiments}
\label{sec:experiments}

We report on English-German NMT. We initialize with a baseline trained on 17.5M sentence pairs from WMT19 news task datasets \cite{barrault-etal-2019-findings}, on which we learn a 32K-merge joint BPE vocabulary \cite{sennrich-etal-2016-neural}.  We validate on newstest2017, and evaluate on newstest2018.

We apply MRT only during fine-tuning, following previous work \cite{edunov2018classical, shen2016minimum}. In early experiments, we found that training from scratch with discriminative objectives (sequence- or document-based) is ineffective. We suspect samples produced early in training are so unlike the references that the model never receives a strong enough signal for effective training.

We fine-tune on old WMT news task test sets (2008-2016) in two settings. With \textbf{random batches}  sentences from different documents are shuffled randomly into mini-batches. In this case doc-MRT metrics are over pseudo-documents. With \textbf{document batches} each batch contains only sentences from one document, and doc-MRT uses true document context. We use the same sampling temperatures and the same risk sharpness factors for both forms of MRT for each experiment.

For Grammatical Error Correction (GEC) we train on sentences from NUCLE \cite{nucle} and Lang-8 Learner English \cite{lang8} with at least one correction, a total of 660K sentences. We  evaluate on the JFLEG \cite{jfleg} and CoNLL 2014 \cite{conll2014} sets. For GEC experiments we use random batching only.

For all models we use a Transformer model \cite{vaswani2017attention} with the `base'  Tensor2Tensor parameters \cite{vaswani-etal-2018-tensor2tensor}.

We train to validation set BLEU convergence on a single GPU. The batch size for baselines and MLE is 4096 tokens. For MRT, where each sentence in the batch is sampled $N$ times, we reduce batch size by $N$ while delaying gradient updates by the same factor to keep the effective batch size constant  \cite{saunders-etal-2018-multi}. 
At inference time we decode using beam size 4. All BLEU scores are for cased, detokenized output, calculated using SacreBLEU  \cite{post-2018-call}.

\subsection{Computation and sample count}
Our proposed document-MRT approach is more complex than sequence-MRT due to the additional score-aggregation and context-sampling steps. In practice we find that the extra computation of ordering and aggregating sequence scores is negligible when compared to the computational cost of sentence sampling, required for all forms of MRT.

Our MRT experiments use $N=8$ random samples per sentence unless otherwise stated. In this we choose the highest $N$ we can practically experiment with, since previous work finds MRT performance increasing steadily with more samples per sentence \cite{shen2016minimum}.

That we see improvements with so few samples is in contrast to previous work which finds BLEU gains only with 20 or more samples per sentence for sequence-MRT \cite{shen2016minimum,  edunov2018classical}. However, we find that document-MRT allows improvements with far fewer samples, perhaps because the aggregation of scores over sentences in a context increases robustness to variation in individual samples.

Relatedly, we find that add-one BLEU smoothing \cite{lin-och-2004-orange} is required for sequence-MRT as in \citet{shen2016minimum}. However we find that doc-MRT can achieve good results without smoothing, perhaps because n-gram precisions are far less likely to be 0 when calculated over a document.

\subsection{MRT for NMT}

\begin{table}[ht!]
\centering
\small

\begin{tabular}{|p{1.5cm}|S[table-format=2.1] S[table-format=2.1]  c c| }
\hline
  \textbf{Model} & \multicolumn{2}{c}{\textbf{Random batches}} & \multicolumn{2}{c|}{\textbf{Document batches}}\\\hline
  Baseline & \multicolumn{4}{c|}{42.7} \\ 
MLE &\multicolumn{2}{c}{40.0} & \multicolumn{2}{c|}{41.0} \\ \hline 
 & {$N=4$}  & {$N=8$} & $N=4$ &$N=8$\\
\hline
Seq-MRT& 42.6  & 43.5 & 42.6 & 43.5  \\ 
Doc-MRT (random) & 41.7$^\ast$ & 43.1$^\ast$  & 43.1 & 43.0 \\ 

Doc-MRT (ordered) & {\textbf{43.4}} & {\textbf{43.7}} & \textbf{43.4} &  \textbf{43.9}\\
\hline
\end{tabular}

\caption{BLEU on en-de after MLE and MRT under $1 - $sBLEU (seq-MRT) and $1 - $doc BLEU (doc-MRT). Results indicated by $^\ast$ are averages over 3 runs with the same settings, which all came within 0.2 BLEU.}\label{tab:results-bleu}. 
\end{table}

\begin{table}[ht!]
\centering
\small
\begin{tabular}{|l p{1cm} p{1.5cm}|}
\hline
  \textbf{Model} &\textbf{Random batches} & \textbf{Document batches}\\\hline
  Baseline & 39.2 & 39.2 \\ 
MLE   &  41.2 & 40.0\\ \hline 
Seq-MRT &39.4 & 40.5\\ 
Doc-MRT (ordered) & \textbf{39.0}  &\textbf{38.9} \\
\hline
\end{tabular}
\caption{TER on en-de after MLE and MRT under sentence-TER (seq-MRT) and doc-TER (doc-MRT). Lower TER is better.}\label{tab:results-ter}
\label{tab:metric-type}
\end{table}

\begin{table*}[t]
\centering
\small
\begin{tabular}{|l|cccc | cccc|}
\hline
 \textbf{Model}& \multicolumn{4}{c|}{\textbf{JFLEG}} & \multicolumn{4}{c|}{\textbf{CONLL2014}}\\
  & {\bf P} & {\bf R} & {\bf M2} & {\bf GLEU} & {\bf P} & {\bf R} & {\bf M2} & {\bf GLEU}  \\\hline

Baseline  & \textbf{67.3}& 38.2 & \textbf{58.4} & 50.4 & \textbf{54.4} &21.8 & 41.9 &67.3\\ 
MLE  & 64.7 & 37.7  & 56.6 & 50.1  &51.4 & 20.9& 39.8& 67.1\\ \hline 
Seq-MRT & 62.7 & 39.1  & 56.0 &50.0   & 52.4 &24.5 & 42.7& 67.1\\ 
Doc-MRT (ordered) & 64.4 & \textbf{41.0}  & 57.8 &\textbf{51.4}   & 53.2& \textbf{24.6}& \textbf{43.2}& \textbf{67.5}\\ \hline

\end{tabular}
\caption{GEC Precision, Recall, M2, and GLEU after MLE and MRT. MRT is under $1-$sentence-GLEU for seq-MRT and $1-$doc-GLEU for doc-MRT. Both MRT schemes uses random batches and random sentence sampling. Higher scores are better for all metrics.}\label{tab:results-gec}
\end{table*}

In Table \ref{tab:results-bleu}, we fine-tune an en-de baseline on documents from past news sets. We compare sentence-BLEU and document-BLEU MRT to fine-tuning with Maximum Likelihood Estimation (MLE). 

MLE fine-tuning degrades the baseline. This suggests the baseline is well-converged, as is desirable for applying MRT \cite{shen2016minimum}. The degradation is smaller with batches containing only sentences from the same document. We connect this to the idea that NMT batches with fewer sentence pairs have ‘noisier’ estimated gradients, harming training \cite{saunders-etal-2018-multi}. We expect batches of sentences from a single document to be similar and therefore give less noisy gradient estimates.

Both seq-MRT and doc-MRT improve over the baseline with random sampling and $N=8$. We also explore MRT at $N=4$, with batch size adjusted as described in section \ref{sec:experiments} for the same effective batch size per update, and with fewer training steps such that the model `sees' a similar proportion of the overall dataset. We do not report beam sampling results as early experiments indicate beam sampling gives similarly poor results for both seq-MRT and doc-MRT. This may be because beam search produces insufficiently diverse samples for this task \cite{freitag-al-onaizan-2017-beam}.

Sequence-MRT gives a 0.8 BLEU gain over the baseline with both batching schemes using $N=8$ samples, but starts to degrade the baseline with $N=4$ samples. With document batches and  $N=8$ Doc-MRT (ordered) outperforms seq-MRT by a further 0.4 BLEU. With $N=4$  doc-MRT (ordered) still achieves a 0.7 BLEU improvement over the baseline, or a 0.8 BLEU improvement over seq-MRT. We suggest therefore that doc-MRT (ordered) may be a computationally more efficient alternative to seq-MRT when large sample counts are not practical.

For contrast with the ordered document sampling approach of Section \ref{sec:doc-sample}, we give results for doc-MRT (random), which uses randomly sampled contexts. This approach falls significantly behind  doc-MRT (ordered) with either batching scheme. Since doc-MRT (random) with random batches is exposed to randomness at the batch construction, sentence sampling and document sampling stages, these results are averages over 3 experimental runs, which gave fairly consistent results ($<$0.2 BLEU range). In general we do find that results with random batches and random ordering are variable and sensitive to batch size and batching scheme.

We interpret these results by considering the effect on the per-sentence cost for the different schemes. We find MRT works well when sample scores are different enough to be discriminated, but suffers if scores are too different. This is in line with the findings of \citet{edunov2018classical} that including the gold reference causes the model to assign low relative probabilities to every other sample. 

Doc-MRT aggregates scores over many samples, while seq-MRT uses individual scores. We believe this explains the stronger performance of doc-MRT for small values of $N$, especially for the ordered document scheme, which ensures scores are still different enough for MRT to discriminate.

Our approach can also be used with document-level metrics that are not intended to be used with individual sentences.  In Table \ref{tab:results-ter} we demonstrate this with TER, which estimates the edit rate required to correct a set of translation hypotheses. Document-TER MRT improves over a strong baseline, although batching scheme has less of an impact here. Notably seq-level MRT does not improve TER over the baseline, indicating TER may be too noisy a metric for use at the sentence level. 

\subsection{MRT for GEC}

Finally, we apply our MRT approach to the GEC GLEU metric \cite{gleu}, an n-gram edit measure typically used at the document level.  Table~\ref{tab:results-gec} shows that document MRT fine-tuning improves GLEU over the baseline, MLE fine-tuning, and a sequence-GLEU MRT formulation. Also notable is the change in M2, which finds the phrase-level edit sequence achieving the highest overlap with the gold-standard \cite{m2}. MLE and sequence-MRT improve recall at a detriment to precision, suggesting over-generation of spurious corrections. Document-MRT likewise improves recall, but with a precision score closer to the baseline for more balanced performance. There is clear indication of a tension between M2 and GLEU: a small increase in GLEU under doc-MRT on CONLL leads to a large
increase in M2, while a large increase in GLEU under doc-MRT on JFLEG leads to a small decrease in M2.

We note that our improvements on JFLEG are similar to the improvements shown by \citet{neural-rl} for neural reinforcement learning with a sequence-GLEU cost metric. However, their results involve N=20 samples and 600k updates, compared to N=8 and 3k updates with our approach.

\section{Conclusions and future work}
We present a novel approach for structured loss training with document-level objective functions. Our approach relies on a procedure for sampling a set of diverse batch-level contexts using N-wise sample ordering. As well as randomly selecting training data, we assess training with mini-batches consisting only of single document contexts. While the scope of this work does not extend to sampling sentences given document context, this would be an interesting direction for future work.

We demonstrate improvements covering three document-level evaluation metrics: BLEU and TER for NMT and GLEU for GEC.  We finish by noting that the original MERT procedure developed for SMT optimised document-level BLEU and with our procedure we reintroduce this to NMT.

\section*{Acknowledgments}
This work was supported by EPSRC grants EP/M508007/1 and EP/N509620/1 and has been performed using resources provided by the Cambridge Tier-2 system operated by the University of Cambridge Research Computing Service\footnote{\url{http://www.hpc.cam.ac.uk}} funded by EPSRC Tier-2 capital grant EP/P020259/1.

\bibliographystyle{acl_natbib}
\bibliography{refs}

\end{document}